\documentclass[journal]{IEEEtran}
\usepackage{cite}
\usepackage{amsmath,amssymb,amsfonts}
\usepackage{algorithmic}
\usepackage{graphicx}
\usepackage{textcomp}

\usepackage[english]{babel}
\usepackage[T1]{fontenc}
\usepackage{epsfig}
\IEEEoverridecommandlockouts
\usepackage{tabu}
\usepackage{float}
\usepackage{amsmath}
\usepackage{amssymb}
\usepackage{amsthm}
\usepackage[keeplastbox]{flushend}
\usepackage{cite}
\usepackage{subfigure}
\usepackage{epstopdf}
\usepackage{multirow}
\usepackage{caption}
\usepackage{mathtools}
\usepackage{xfrac}
\usepackage{units}
\usepackage[hyphens]{url}
\usepackage{array}
\newcommand{\beql}[1]{\begin{equation}\label{#1}}
\newcommand{\eeq}{\end{equation}}

\newcommand{\be}{\begin{equation}}
\newcommand{\ee}{\end{equation}}
\newcommand{\ba}{\begin{array}}
\newcommand{\ea}{\end{array}}

\ifCLASSINFOpdf
\else
\fi
\begin{document}
%

\title{Benchmarking Inference Performance of Deep Learning Models on Analog Devices}
%
%
%
\author{Omobayode~Fagbohungbe,~\IEEEmembership{Student~Member,~IEEE,}
        ~Lijun~Qian,~\IEEEmembership{Senior~Member,~IEEE}
\thanks{The authors are with the CREDIT Center and the Department
of Electrical and Computer Engineering, Prairie View A\&M University, Texas A\&M University System, Prairie View, 
TX 77446, USA. Corresponding author: Omobayode Fagbohungbe e-mail: ofagbohungbe@student.pvamu.edu}}
\maketitle

\begin{abstract}
Analog hardware implemented deep learning models are promising for computation and energy constrained systems such as edge computing devices. However, the analog nature of the device and the associated many noise sources will cause changes to the value of the weights in the trained deep learning models deployed on such devices. In this study, systematic evaluation of the inference performance of trained popular deep learning models for image classification deployed on analog devices has been carried out, where additive white Gaussian noise has been added to the weights of the trained models during inference. It is observed that deeper models and models with more redundancy in design such as VGG are more robust to the noise in general. However, the performance is also affected by the design philosophy of the model, the detailed structure of the model, the exact machine learning task, as well as the datasets.
\end{abstract}


%
\IEEEpeerreviewmaketitle

\section{Introduction}
\label{sec:Introduction}
Deep Learning (DL) has enjoyed and continues to enjoy renewed interest from researchers due to their ability to achieve or even surpass human-level performance for many cognitive applications. The resurgence can be attributed to large scale dataset, high-performance hardware, new activation functions and more sophisticated optimization methods~\cite{noh2017regularizing}. These factors have led to an advancement in the field of deep learning favoring the design of deep, large, and very complex models.
Despite their unprecedented level of performance and improvement in their design in recent years, DL models require high computational and energy resources during training and inference~\cite{mixedsignal,Li}. The high computational resource requirement is because of the intense fundamental operations by these models, such as dot product of vector and matrix, and multiplications of matrices, during training and inference~\cite{xiao,charan}. This is further complicated by the large  increase in the quantity of these operations with the increase in the size of the models. The high energy requirement can be attributed to the data intensive nature of the models and the huge memory and memory bandwidth requirement of the deep and large models. As a result, typical CPU and GPU on edge devices do not have enough memory resource to process these models efficiently as the CMOS technology is approaching its limit and making it a performance bottleneck~\cite{mixedsignal,xiao}.  Hence, models are stored in the off-chip memory leading to the presence of a memory wall, a physical separation between the processing unit and the memory~\cite{xiao}. This separation means there is a need for constant shuttling for data access between the processing unit and the off-chip memory that leads to high energy consumption and high latency~\cite{joshi}. 
 \begin{figure}[htbp]
	 \centering
    	 \includegraphics[width=8.6cm]{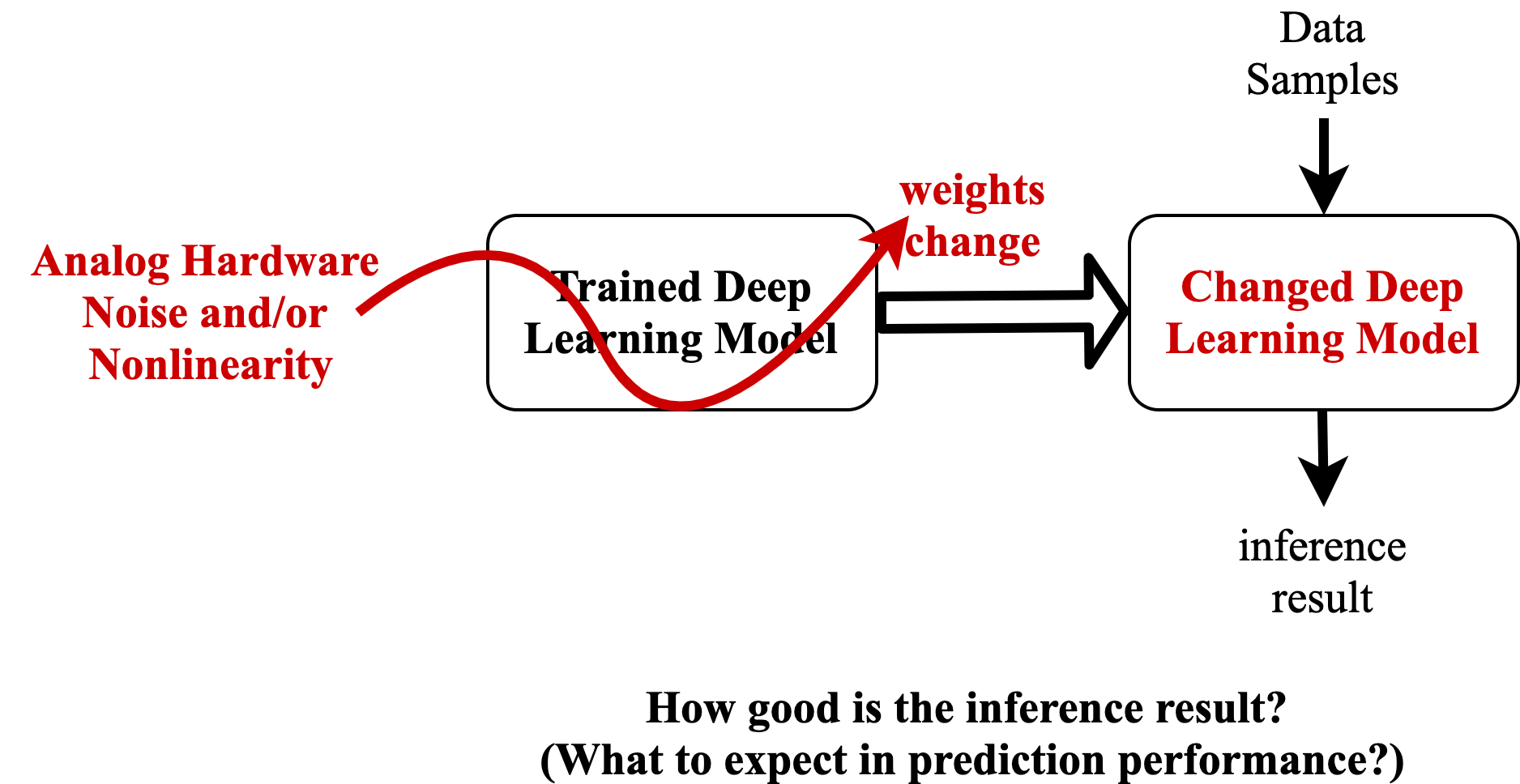}
     	\caption{Noise induced performance degradation of hardware implemented deep learning models }
    \label{fig:Hardwareerrorinducedperformancedegradation}
\end{figure}

\begin{table*}
\centering
 \caption{The details of the DL models and dataset used in the experiments. The baseline inference accuracy is obtained by performing inference on the test set of the dataset when noise has NOT been added to the weights of the models.}
\label{table:details}
 \begin{tabular}{|*{6}{c|} }
 \hline
 Model Name & Dataset & Number of Classes & Model Input Dimension & Baseline Inference Accuracy (\%)\\  
 \hline\hline
 ResNet\_18 & Imagenet & 1000 & 224*224*3 & 89.50\%   \\ 
  \hline
 ResNet\_34 & Imagenet& 1000 & 224*224*3 & 91.44\% \\ 
   \hline
 ResNet\_50 & Imagenet& 1000 & 224*224*3 & 93.15\%  \\
   \hline
ResNet\_50\-SE & Imagenet& 1000 & 224*224*3 & 93.76\%  \\
 \hline
 Shufflenet\_V2\_1.0X & Imagenet& 1000 & 224*224*3 & 88.69\% \\ 
  \hline
Shufflenet\_V2\_0.5X & Imagenet& 1000 & 224*224*3 & 82.37\% \\ 
  \hline
 VGG\_16 & Imagenet& 1000 & 224*224*3 & 90.17\% \\  
   \hline
 ResNet\_20 & CIFAR10& 10 & 32*32*3 & 89.50\%   \\ 
  \hline
ResNet\_32 & CIFAR10 & 10 & 32*32*3 & 91.44\% \\ 
   \hline
ResNet\_56 & CIFAR10& 10 & 32*32*3 & 93.15\%  \\
   \hline
\end{tabular}
\end{table*}

At the same time, there has been increasing demand for high performance and energy efficient computing in recent years. This demand has been further fueled by the desire to deploy DL models on Internet-of-Things (IoT) and edge devices which operate within a tight computational resource and power envelope~\cite{NoisyNN}. This tight requirement and the desire to fix compute and memory transfer bottlenecks in current set of hardware has led to a significant interest in analog specialized hardware for DL, as they have the potential to deliver at least 2X better performance than the conventional digital hardware in both speed and energy efficiency~\cite{Ni,Shen_2017}. In fact, they can deliver at projected throughput of multiple tera-operations (TOPs) per seconds and also achieve femto-joule energy budgets per multiply-and-accumulate (MAC) operation~\cite{charan,Bennett2020,Burr,marinella}. The improvement can be attributed to the use of non-volatile memory cross bar arrays to encode DL model weights and biases, a form of computing known as in-memory computing. The arrays have a multi-level storage capability and also allow a single time step matrix-vector multiplication based on Kirchhoff's circuit laws~\cite{NoisyNN,xiao,MITTAL2020101689,joshi}. 
Despite these advantages, analog accelerators do not have the bit-exact precision enjoyed by digital hardware and suffer from noise~\cite{NoisyNN}, as shown in Figure~\ref{fig:Hardwareerrorinducedperformancedegradation}. 
This can be attributed to many factors such as thermal noise, quantization noise, circuit non-linearity, and device failure~\cite{mixedsignal}. These disadvantages can affect the reliability of DL models in the form of performance degradation depending on the prevailing factors. Although there are some studies on the robustness of DL models 
to label noise~\cite{rolnick2017deep}, very little is known on the inference behavior of trained DL models when the weights are subject to analog noises.  

The objective of this paper is to systematically  study the effect of additive noise on the performance of trained DL models implemented in analog accelerators. Specifically, we try to establish the benchmark of performance degradation and provide observations on the detailed behavior of some popular computer vision DL models for image classification task in the presence of analog noise induced weight changes, as illustrated in Figure~\ref{fig:Hardwareerrorinducedperformancedegradation}. The noise is modeled as  additive white noise to the weights of trained DL models implemented in analog accelerators. We vary the strength of the noise and the impact of the noise on each layer of the model is investigated. The remainder of this paper is organized as follows: The methodology used for this work is discussed in Section~\ref{sec:methods}. Experimental results and analysis are given in Section~\ref{sec:result}. Further discussions and related works are reviewed in Section~\ref{sec:discussion}.  Section~\ref{sec:conclusion} concludes the paper.

\section{Methodology}
\label{sec:methods}
The effect of additive white noise on the performance of  pre-trained DL models for computer vision task during inferencing is investigated. The DL models used for this work are selected models used for image classification task. The performance metrics under observation is the percentage of classification accuracy (100\% - error) of the model on the test data. 

The noise was modeled as a white Gaussian noise of zero mean and a standard deviation of $\sigma_{noise}$. The standard deviation of a white noise can be interpreted as the energy of the noise. The value of $\sigma_{noise}$ was decided using a term known as the signal to noise ratio (SNR). The SNR in this context is defined as the ratio of the standard deviation of the weights in a layer $\sigma_w$, to the standard deviation of the noise added to that layer $\sigma_{noise}$. This is defined mathematically as:
\begin{equation}
SNR=\frac{\sigma_w}{\sigma_{noise}}
\end{equation}
A Gaussian noise of zero mean and standard deviations equivalent to 1\%, 10\%, 20\%, 40\%, 60\% and 100\% of the standard deviation of the weights of a particular layer $\sigma_w$, is added to the same layer. This is equivalent to the SNR of 100, 10, 5, 2.5, 1.67 and 1, respectively.

The pre-trained model of interest was put in inference mode and performance measurement on the test data is taken to establish the baseline. Thereafter, the effect of the additive noise was investigated by adding Gaussian noise of the desired standard deviation or energy to the first layer of the model. The model was then set to inference mode and the performance measurement was taken. This was done multiple times and the average of the testing accuracy was obtained. This process was then repeated for the rest of the layers and the testing accuracy was recorded. The classification accuracy due to the present of noise $a_i$ in the model is then normalized with the baseline classification accuracy $a_o$, that is given by:
 \begin{equation}
A_i=\frac{a_i}{a_o}
\end{equation}
where $A_i$ is the normalized classification accuracy due to the present of noise in layer $i$. 

\begin{table*}
\centering
\caption{The detailed overview of the experiment design for this work. }
\label{table:experimental_design}
\begin{tabular}{ |c|c|c|c| } 
\hline
Experiment Design & Aim of Experiment & Model and dataset Used \\
\hline
\multirow{3}{4em}{1} & To establish the performance degradation & (i) VGG16, ShuffleNet\_V2\_R1 and ResNet50\\ 
& suffered by deep learning models due to  analog noise &  models using Imagenet dataset \\
& in all their layers &  \\ 
\hline

\multirow{4}{4em}{2} & (i) To establish general performance degradation trends & (i) ResNet50 on Imagenet dataset\\ 
& by deep learning models due to  analog noise in a single layer &  \\
& (ii) To establish general performance degradation trends & (ii) ResNet50 on Imagenet dataset  \\ 
& by deep learning models due to  analog noise in some layers &  \\
\hline

\multirow{3}{4em}{3} & To investigate the effect of model depth & (i) ResNet 18,34 and 50 on Imagenet dataset\\ 
& on the degradation of deep learning models due to  presence of  & (ii) Modified ResNet 20,32 and 56 on CIFAR10 dataset \\ 
&  analog noise in a single layer &  \\ 
\hline

\multirow{3}{4em}{4} & To investigate the effect of model design philosophy & (i)VGG16 vs ShuffleNet\_V2\_R1 vs ResNet50 \\ 
&   on the degradation of deep learning models due to  presence of  & on Imagenet dataset \\
& analog noise in a single layer & (ii) ResNet50 vs Resnet50-SE on Imagenet dataset \\ 
\hline

\multirow{3}{4em}{5} & To investigate the effect of model compression & (i)ShuffleNet\_V2\_R1 vs ShuffleNet\_V2\_R0.5 \\ 
&   on the degradation of deep learning models due to  presence of  & on Imagenet dataset \\
& analog noise in a single layer &  \\ 
\hline

\end{tabular}
\end{table*}

\section{Experimental Results and Analysis}
\label{sec:result}
\subsection{Experimental Setup}
\label{subsec:ExpSetup}
A series of experiments have been conducted where the CIFAR10 and the Imagenet datasets were used. The overview of the experiments carried out is stated in Table \ref{table:experimental_design}. The details of the dataset used such as size of the images and the number of classes are given in Table~\ref{table:details}. The CIFAR10 dataset are tiny images which are mutually exclusive and has no semantic overlaps between images of different classes. The Imagenet dataset, which is bigger than the CIFAR10 dataset in dimension and size, is originally part of the ILSVRC 2012 dataset. It contains classes which are either internal or leaf nodes but do not overlap. 

The details of the DL models used for this work is provided in Table \ref{table:details}. The pre-trained models used in this work were chosen based on popularity and also because the models obtained good baseline results on the classification tasks. The baseline results of each of the models are also shown in Table \ref{table:details}.
The DL models tested on the Imagenet dataset were obtained from the tensorpack model zoo. Tensorpack is a neural network interface based on Tensorflow~\cite{tensorpack}. The models are ResNet models~\cite{resnet,Resnetse}, Shufflenet models~\cite{shufflenet} and VGG16 model~\cite{vgg}. The models tested on the CIFAR10 dataset were trained and tested following the algorithm in~\cite{towards}. The models are modified version of ResNet models suited for the CIFAR10 dataset. This was done in Keras deep learning framework using Tensorflow backend. All the experiments were conducted using a NVIDIA Tesla V100-DGXS-32GB GPU.

\subsection{Results Analysis With additive Noise on All the Weights of the Deep Learning Model}
\begin{figure}[htbp]
	 \centering
    	 \includegraphics[width=8cm]{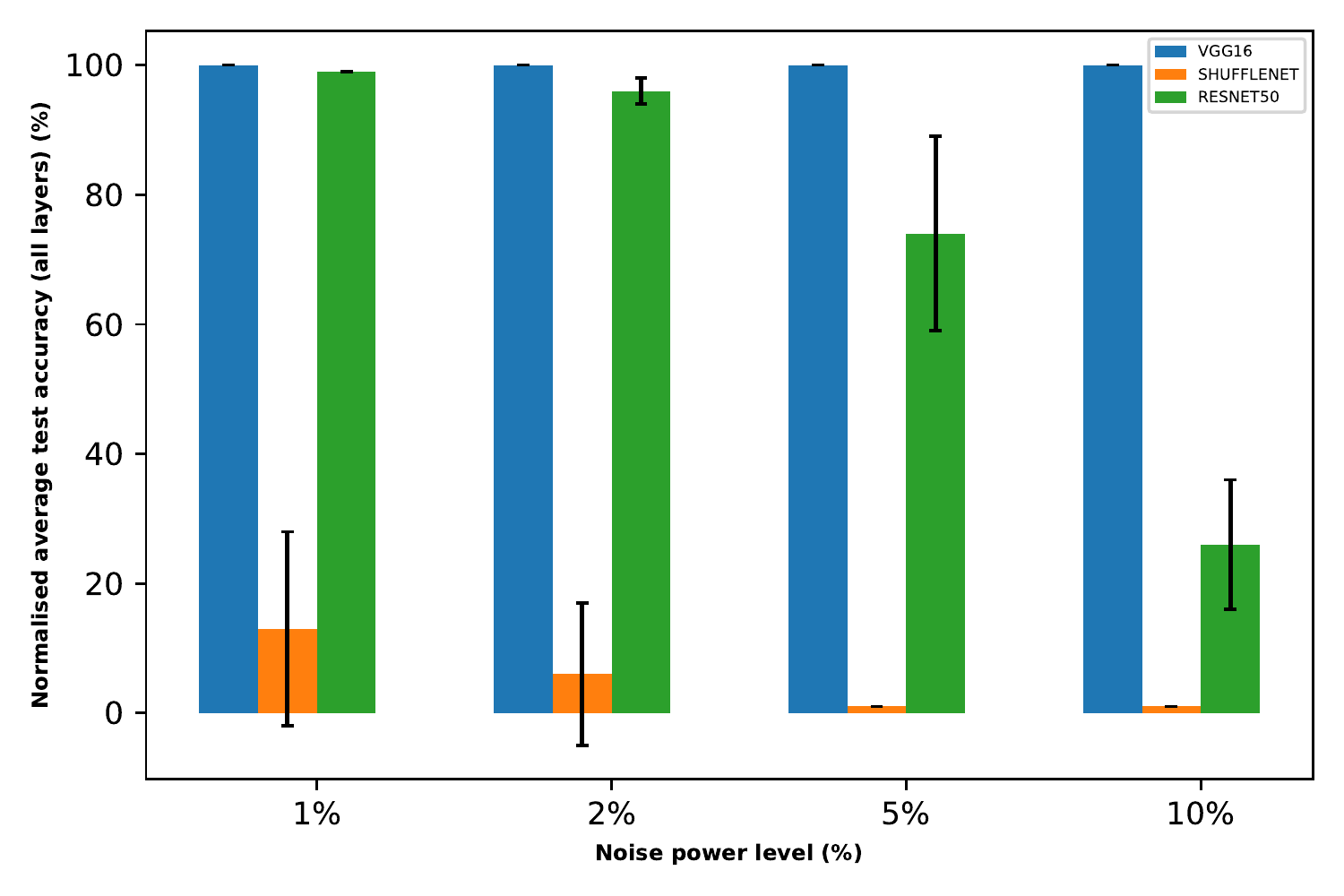}
     	\caption{The inference performance of pre-trained VGG16, ResNet50, and Shufflenet when noise added to all the weights }
    \label{fig:VGG16ResNet50ShufflenetAllWeightsWithNoise}
\end{figure}

In this section, the effect of additive noise on three popular deep learning models for image classification, namely, VGG16, ResNet50, and Shufflenet, has been evaluated. Specifically, the three models are pre-trained on Imagenet dataset. Then their inference performance on the test dataset is measured when additive white Gaussian noise with certain energy level (1\%, 2\%, 5\% and 10\%) has been added to all the weights of the models. The resulted prediction accuracy of the three models are shown in Figure~\ref{fig:VGG16ResNet50ShufflenetAllWeightsWithNoise} and Table~\ref{tab:all_model_comparision}. It should be noted that while Table~\ref{tab:all_model_comparision} contains the absolute value for the classification accuracy for the various models, Figure~\ref{fig:VGG16ResNet50ShufflenetAllWeightsWithNoise} contains the values of the classification accuracy for each noise power when normalised with the corresponding classification accuracy for each model type in the absence of noise (noise power of 0\%).
 
 It can be observed that VGG16 has the best performance and Shufflenet model has the poorest performance. The VGG16 model did not suffer any degradation in performance when additive white Gaussian noise of 10\% energy level  was added, attesting to the noise resistant behaviour of the model. However, Shufflenet model suffered huge degradation in performance even for  1\% energy level white Gaussian noise. The model suffered 87\%,94\%,99\% and 99\% in degradation for a noise energy levels of 1\%,2\%,5\% and 10\% respectively. Although the ResNet50 model also suffered some performance degradation due to the noise, the degradation in performance is not as severe as the degradation observed in the Shufflenet model. 
 
 The difference in noise resistant ability of the various DL models can be attributed to the number of parameters in the model~\cite{SIETSMA199167}. There are about 2.2 million , 25.6 million and 138.4 million parameters in Shufflenet, ResNet50 and VGG16 models parameter. Models with very large parameters have their learning ability shared among many parameters. Hence, they are more resistant to noise that models with fewer parameters.

\subsection{Layer-by-Layer Analysis of the Deep Learning Models With Noise added Per Layer }
The previous section gives the inference performance of three popular deep learning models when additive white Gaussian noise is added to all the weights in the models. This provides a general measure of the noise resistance ability of the models. In this section, we go a step further to examine the models layer-by-layer. The goal is to evaluate how each layer in a deep learning model affects the inference performance when exposed to noise. The observation will help us understand  the role of the layers and their respective noise resistance property, and eventually be able to design more robust deep learning models for noisy analog devices.
\begin{table}[]
\centering
\caption{Comparison of the performance of pre-trained VGG16, Shufflenet\_V2\_R\_1.0X and ResNet50 Models in the presence of noise in all its layer during inference when tested with Imagenet dataset. The performance metric is the model classification accuracy}
\label{tab:all_model_comparision}
\begin{tabular} {{|c|c|c|c|c|c|c|c|c|c|}}
     \hline 
    & \multicolumn{4}{c}{Noise power (\%)} & \\ \hline\hline
     Model\_Name & 0\% & 1\% & 2\% & 5\% & 10\% \\ \hline 
     VGG16  & 90\% & 90\% & 90\% & 90\% & 90\%\\ \hline
     Shufflenet\_V2\_R1.0X & 89\% & 12\% & 6\% & 1\% & 1\% \\ \hline
     ResNet50 & 93\% & 92\% & 89\% & 69\% & 24\%  \\ \hline
\end{tabular}
\end{table}

\subsubsection{Performance Analysis of pre-trained ResNet50}
\begin{figure}[]
	 \centering
    	 \includegraphics[width=8cm]{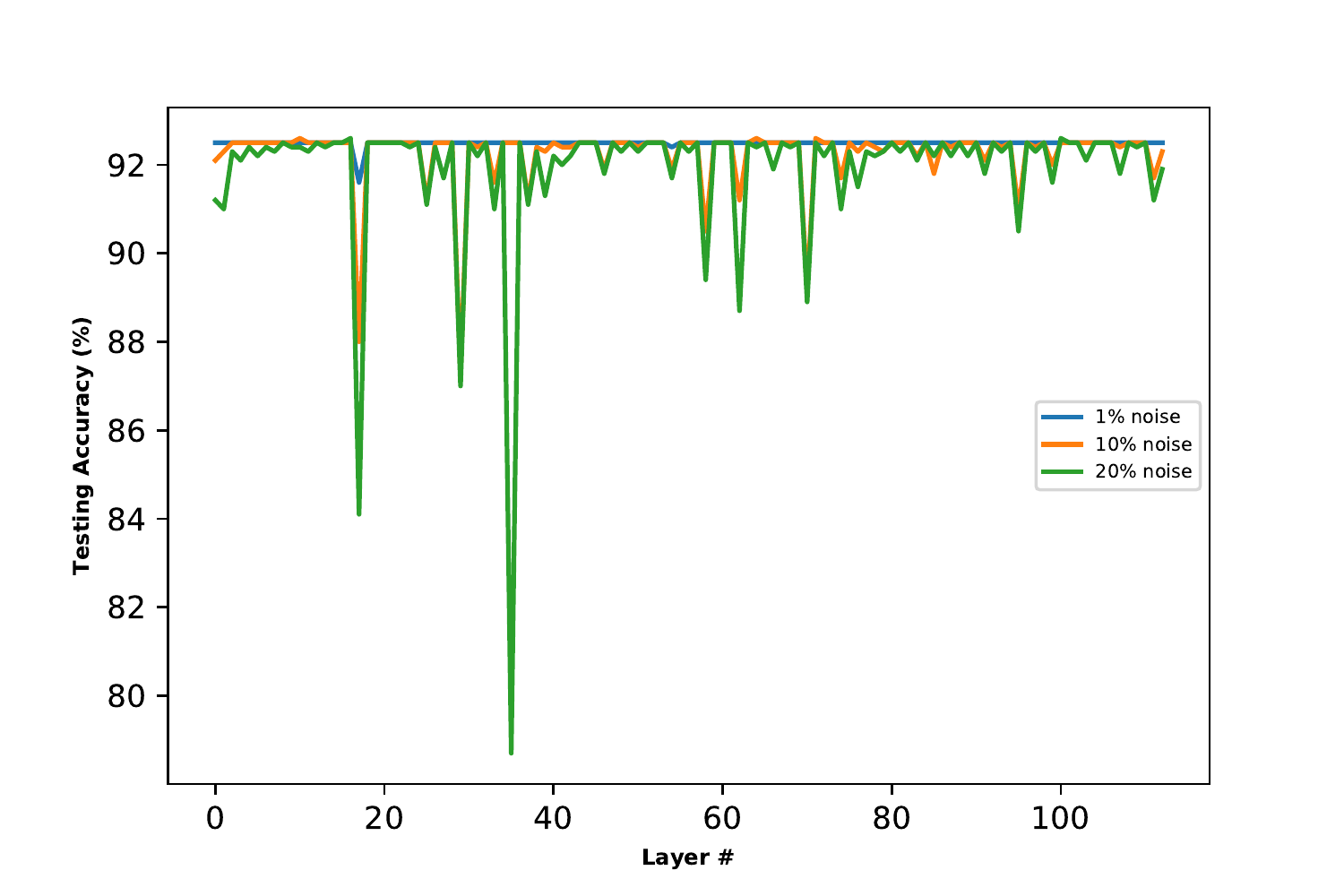}
     	\caption{The inference performance of pre-trained ResNet50 trained on Imagenet dataset in the presence of noise (seed=40) in its layer (one layer at a time) }
    \label{fig:ResNet_2}
\end{figure}

\begin{figure}[]
	 \centering
    	 \includegraphics[width=8cm]{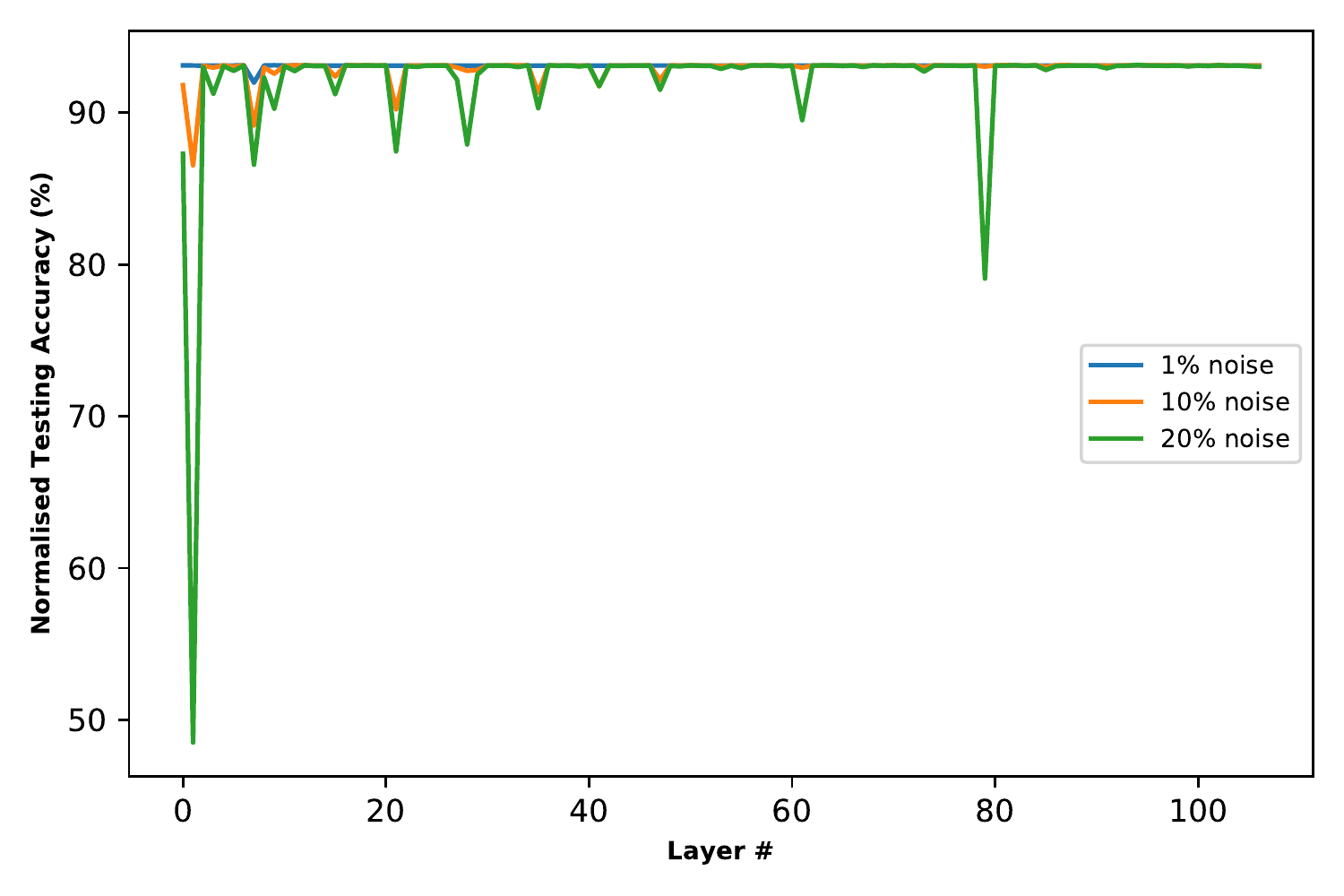}
     	\caption{The inference performance of pre-trained ResNet50 trained on Imagenet dataset in the presence of noise (seed=30) in its layer (one layer at a time) }
    \label{fig:ResNet_seed}
\end{figure}

The effect of additive white Gaussian noise of various power levels on the parameters (trainable and untrainable) of each layer of ResNet50 model trained on Imagenet dataset are shown in Figure \ref{fig:ResNet_2} and \ref{fig:ResNet_seed} (with different seeds of the pseudorandom noise). In this work, the batch normalization parameters were considered to be a single layer. This is because their properties are a little different from the properties of the weight and bias of a neural network. It was observed that the neural network showed a certain degree of resistance to the noise at lower power level for all the layers in the model. It was also observed that the level of resistance reduces as the noise energy level increases. For noise energy level that the model didn't show strong resistance to the effect of additive white noise, it was observed that the later layers are more resistance to the noise than the layers that are earlier in the network. This trend was also observed for other models irrespective of the type of model used or the dataset they were trained on. However, for each specific layer, although it appears that there is some pattern in the degradation of performance, it is still quite random for individual layers in the model. For instance, when the energy of the noise is the same but  the seed of the pseudorandom noise changes, the pattern of the degradation is very different as shown in Figure~\ref{fig:ResNet_2} with seed=40 and Figure~\ref{fig:ResNet_seed} with seed=30.

\begin{figure}[]
	 \centering
    	 \includegraphics[width=8cm, height=5.85cm]{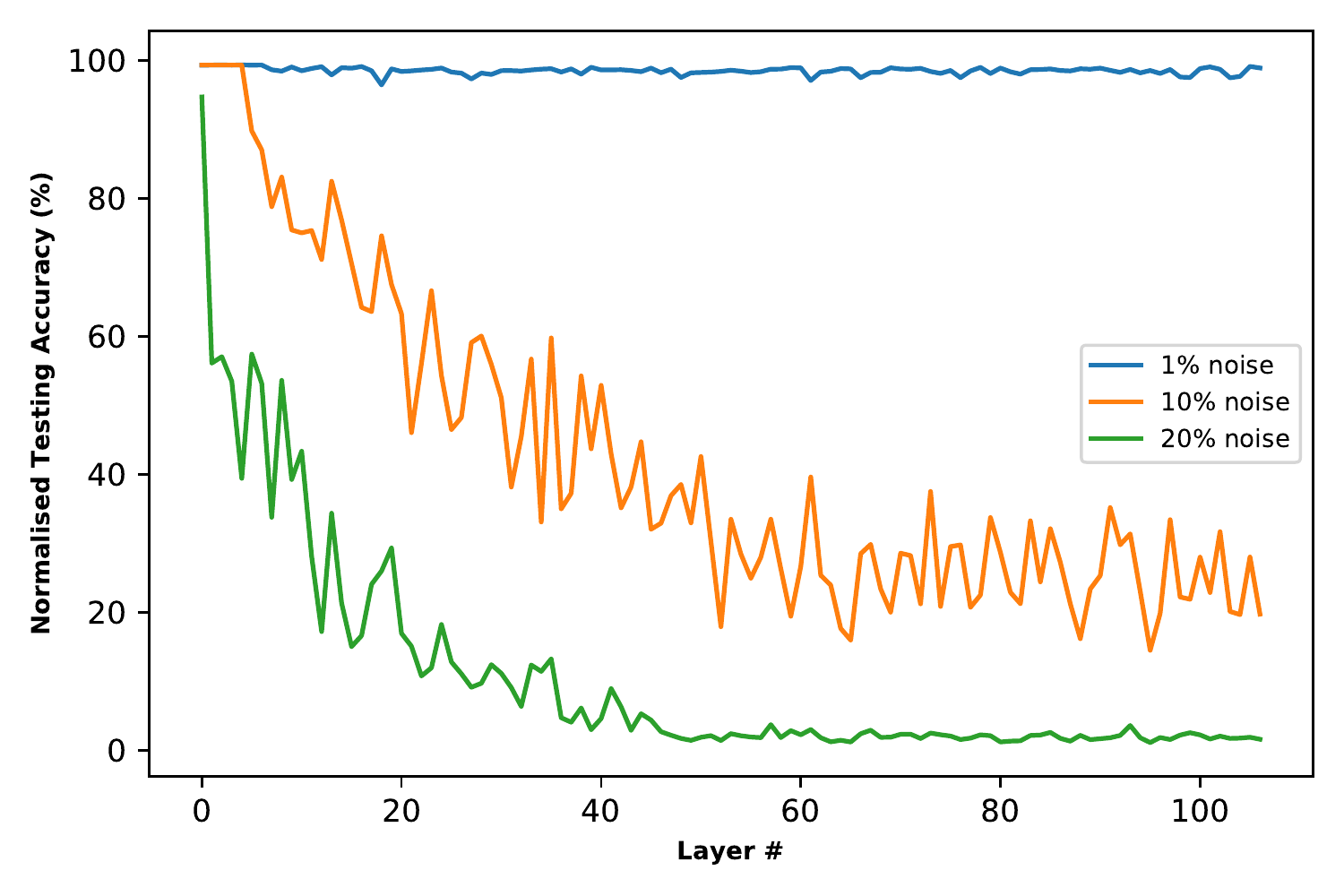}
     	\caption{The inference performance of pre-trained ResNet50 trained on Imagenet dataset in the presence of noise in multiple layers (layer 1 to  layer $L$, $L$ is the layer index).}
    \label{fig:ResNet_1}
\end{figure}

Figure \ref{fig:ResNet_1} shows the effect of adding additive white noise to more than one layer in the network starting with layer one. In other words, the performance result at  layer $L$ means that noise of a particular power level was added to all the layers from layer 1 to  layer $L$. For a fixed noise level, it was observed that the amount of degradation in model classification accuracy increases as we increase the number of layer until the model testing accuracy approach zero. It can also be observed from the figure that the depth of the layer at which the model classification accuracy reduces to almost zero reduces as the energy level of the noise increases. 

In order to aid better summarization and analysis of results, a new metric called average normalized percentage classification accuracy per layer is introduced. This is mathematically defined as:
\begin{equation}
A_{avr}=\frac{\sum\limits_{i=1}^{N} A_i}{N}
\end{equation}
where $A_{avr}$ is the average normalized percentage classification accuracy per layer.
This new metrics is a metric of a model rather than the metric of a layer $l_i$ in a model. This metric is different from $A_i$ which is a metric of a particular layer in a model. The comparisons of the various models are then done in the subsections below.

\subsubsection{Analysis of ResNet models trained on Imagenet and CIFAR10 datasets}
\label{subsec:caseA}
In this section, the analysis of the performance of various ResNet models trained on Imagenet and CIFAR10 dataset for classification task is given, when additive white Gaussian noise at a particular energy level is added to the various layers of the models. For the Imagenet dataset, the models of interest are ResNet18, ResNet34 and ResNet50 models. The models are the same in design principles except in depth with ResNet18 being the shallowest and ResNet50 being the deepest. The models of interest for the CIFAR10 datasets are ResNet20, ResNet32 and ResNet56 models with ResNet20 being the shallowest and ResNet56 being the deepest. These models are a variant of the original ResNet18, ResNet34 and ResNet50 models, respectively. The significant difference in the 2 category of models lies in the fact that the model architecture has been modified to be suited for use in the CIFAR10 dataset but the design philosophy is essentially the same.

\begin{figure}[]
  \includegraphics[width=8cm, height=5.85cm]{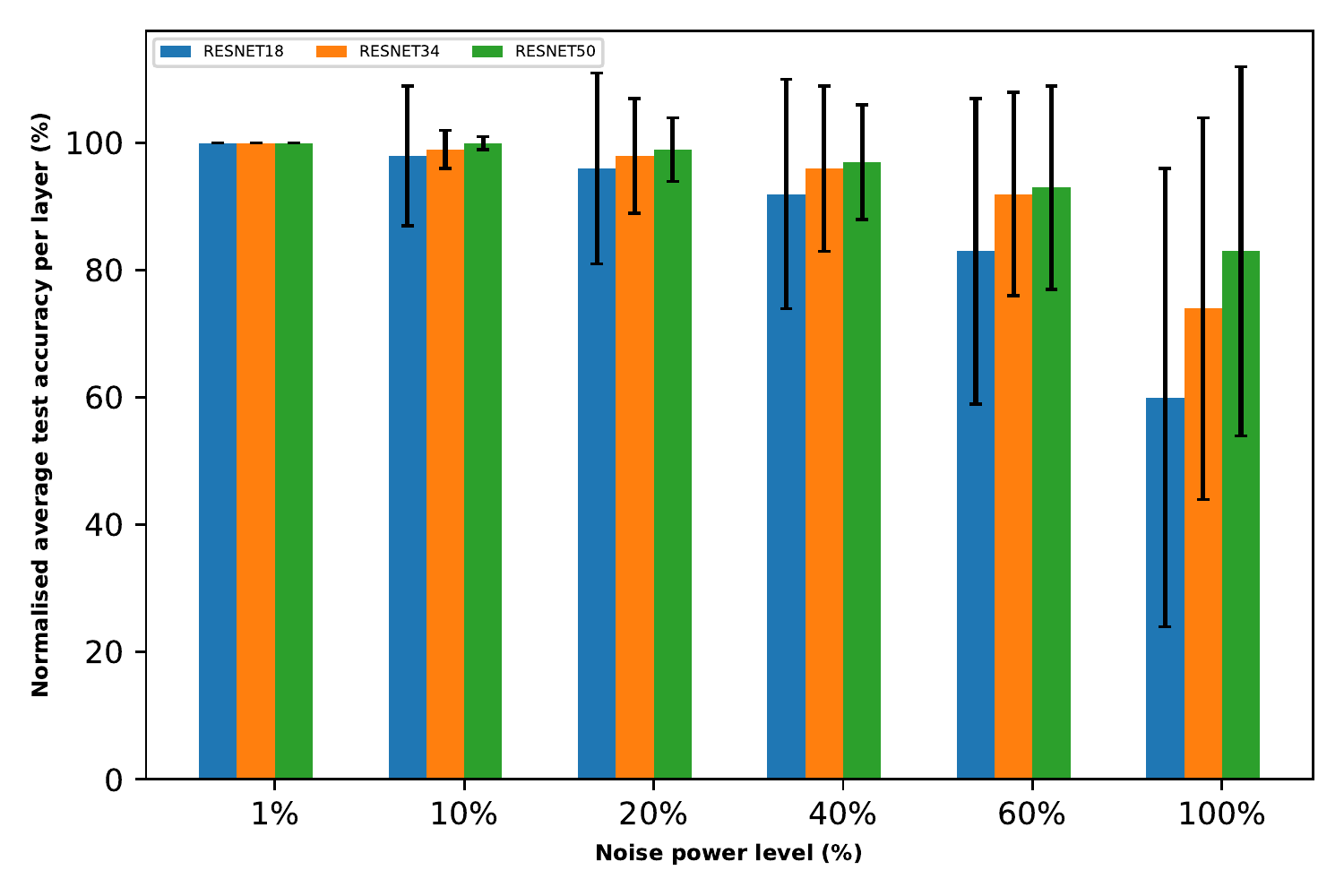}
\caption{Comparison of the performance of pre-trained ResNet18, ResNet34 and ResNet50 in the presence of noise in its layer during inference when tested with Imagenet dataset}
\label{fig:ResNet18_34_50}
\centering
\captionof{table}{Details of the ResNet18, ResNet34 and ResNet50 models. The batch normalization parameter is considered as a layer. The number of parameter includes both trainable and non-trainable parameters}
\label{tab:ResNet18_34_50}
\begin{tabular}{ | c | c | c | c |}
     \hline
     Model\_Name & \# of Layers & \# of Parameters \\ \hline \hline	
     ResNet18  & 41 & 11,699,112\\ \hline
     ResNet34 & 73 & 21,814,696 \\ \hline
     ResNet50 & 107 & 25,610,152 \\ \hline
\end{tabular}
\end{figure}

\begin{figure}[]
  \includegraphics[width=8cm]{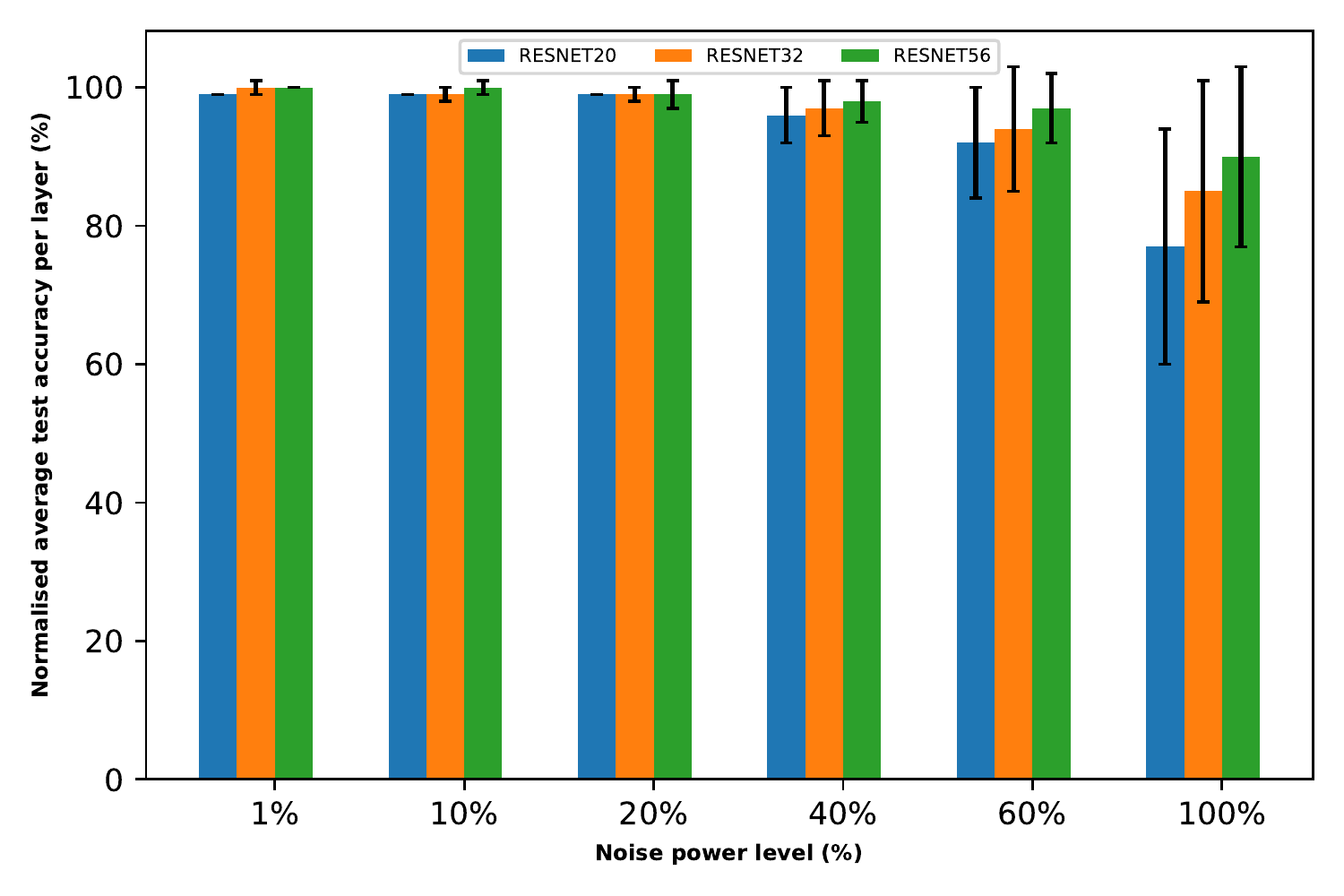}  
\caption{Comparison of the performance of pre-trained ResNet20, ResNet32 and ResNet56 in the presence of noise in its layer during inference when tested with CIFAR10 dataset}
\label{fig:ResNet20_34_56}
\end{figure}

The performance comparison of the ResNet model trained on Imagenet and CIFAR10 datasets are shown in Figure \ref{fig:ResNet18_34_50} and \ref{fig:ResNet20_34_56}, respectively. It is clear that the lowest degradation in performance (100-$A_{avg}$) was observed for each of the model at lower noise power, as expected. This observations buttresses the noise resistant ability of deep neural networks. It was shown that the amount of degradation also increases with the increase in the noise power level for each model. Furthermore, It was observed that models with more depth suffers less degradation as compared to models with less depth. It was also observed that the models trained on Imagenet dataset suffers more degradation in performance when compared with the equivalent model for the CIFAR10 dataset. This may be due to the fact that the classification task for the Imagenet dataset is more complex than that in the CIFAR10 dataset.

\begin{table}[]
\centering
\caption{Comparison of the performance of pre-trained ResNet20, ResNet32 and ResNet56 in the presence of noise in its layer during inference when tested with CIFAR10 dataset}
\label{tab:ResNet20_34_56}
\begin{tabular}{ | c | c | c | c |}
     \hline
     Model\_Name & \# of Layers & \# of Parameters \\ \hline \hline	
     ResNet20  & 40 & 274,442\\ \hline
     ResNet32 & 65 & 470,218 \\ \hline
     ResNet56 & 113 & 861,770 \\ \hline
\end{tabular}
\end{table}

\begin{figure}
  \includegraphics[width=8cm,height=5.85cm]{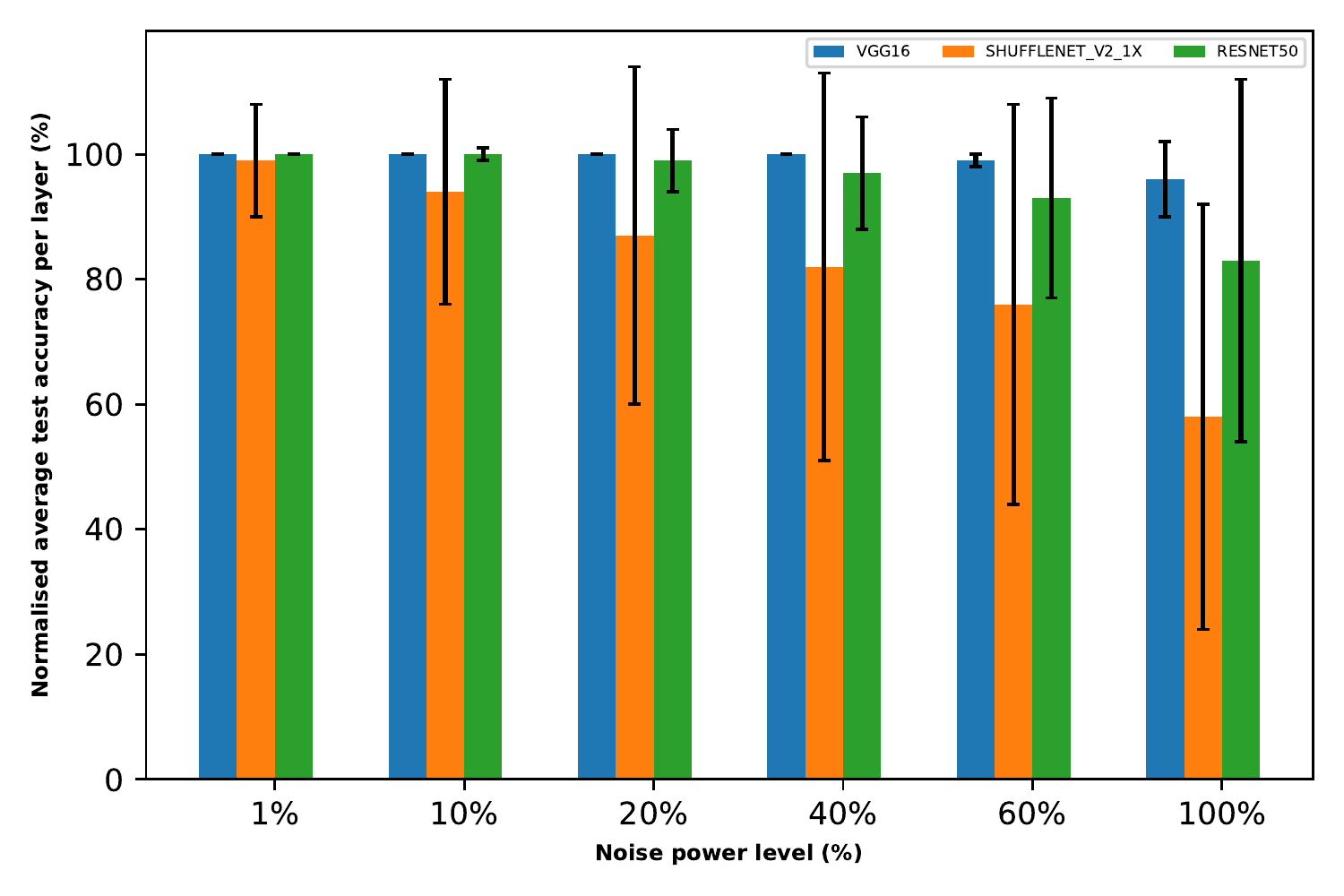}
\caption{Comparison of the performance of pre-trained VGG16, Shufflenet\_V2\_1X and ResNet50 in the presence of noise in its layer during inference when tested with Imagenet dataset}
\label{fig:VGG16_SHUFF_ResNet}
\end{figure}
\begin{table}
\centering
\caption{Details of the VGG16, Shufflenet\_V2\_1X and ResNet50 models. The batch normalization parameter is considered as a layer. The number of parameter includes both trainable and non-trainable parameters}
\label{tab:VGG16_SHUFF_ResNet}
\begin{tabular}{ | c | c | c | c |}
     \hline
     Model\_Name & \# of Layers & \# of Parameters \\ \hline \hline	
     VGG16  & 16 & 138,357,544\\ \hline
     Shufflenet\_V2\_1X & 113 & 2,295,940 \\ \hline
     ResNet50 & 107 & 25,610,152 \\ \hline
\end{tabular}
\end{table}




\begin{figure}
  \includegraphics[height=5.85cm, width=8cm]{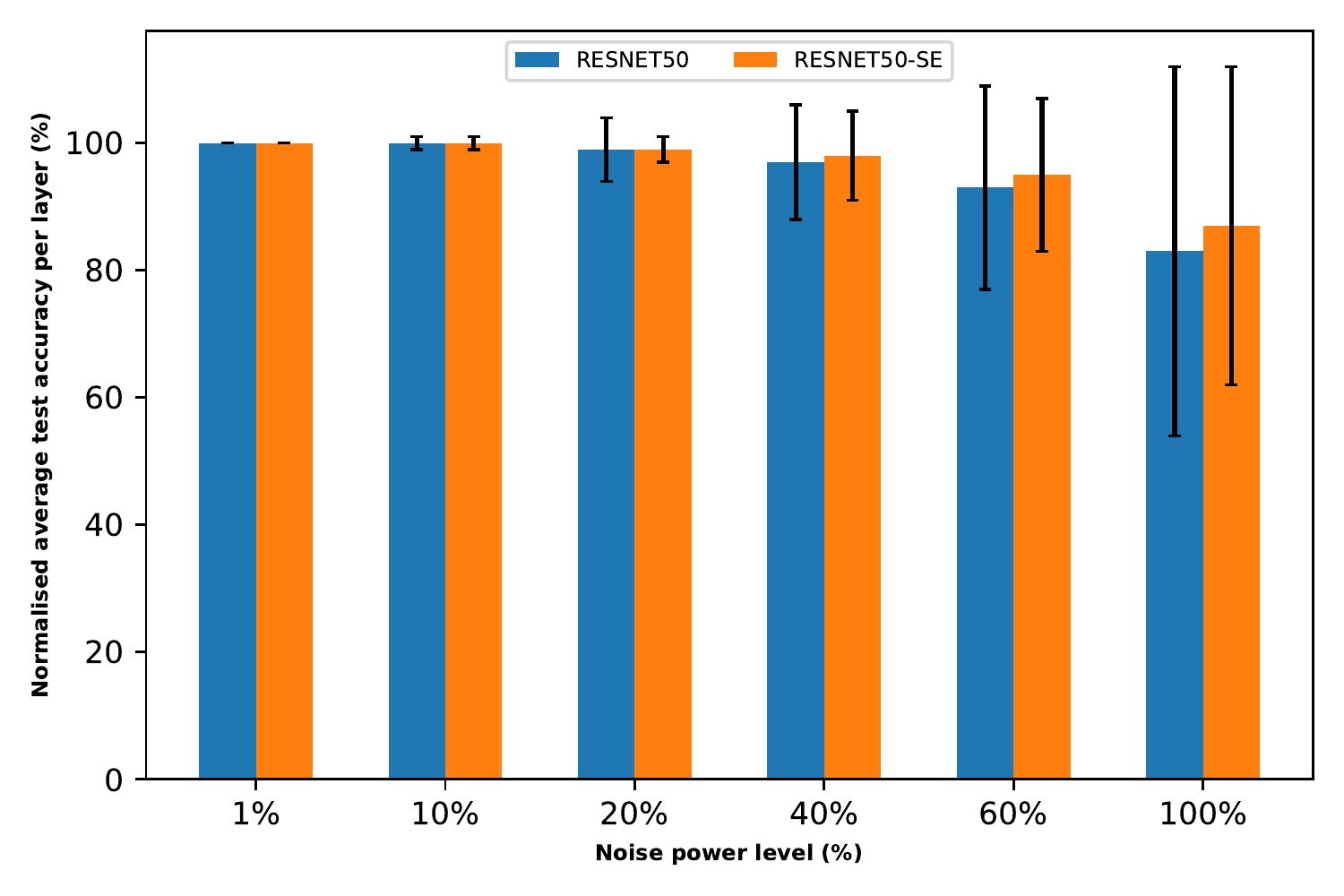}
\caption{Comparison of the performance of pre-trained ResNet50 and ResNet50-SE in the presence of noise in its layer during inference when tested with Imagenet dataset}
\label{fig:ResNet50_se}
\end{figure}

\begin{table}
\centering
\captionof{table}{Details of the ResNet50 and ResNet50-SE models. The batch normalization parameters are grouped together and considered as a layer. The number of parameter includes both trainable and non-trainable parameters}
\label{tab:ResNet50_se}
\begin{tabular}{ | c | c | c | c |}
     \hline
     Model\_Name & \# of Layers & \# of Parameters \\ \hline \hline	
     
     ResNet50 & 107 & 25,610,152 \\ \hline
     ResNet50-SE  & 139 & 28,141,144\\ \hline
\end{tabular}
\end{table}

\subsubsection{Analysis of VGG and Shufflenet models trained on Imagenet}
\label{subsec:caseB}
The effect of model design principles on the ability of neural network to resist additive white Gaussian noise of different power level is investigated in this section. This is done by comparing the performance of three different models, trained on Imagenet dataset but with different architecture design principles, when additive white Gaussian noise of various power is injected into their layers. This is different from the case in Section~\ref{subsec:caseA} where the models were the same in design principles but vary in depth. The models investigated and compared in this section are VGG16, ResNet50 and Shufflenet (Version 2, ratio 1.0x) and ResNet50-SE models. 

Figure \ref{fig:VGG16_SHUFF_ResNet} shows the performance comparison between VGG16, Shufflenet\_V2\_R1.0X and ResNet50 when noise is injected into their layers. The trend in the performance of each of the model were very similar to the one observed in section~\ref{subsec:caseB} although the degradation in performance differs from one model to the other. The VGG16 model has the best noise resistance property among the three models while Shufflenet\_V2\_R1.0X having the worse performance. In fact, the VGG16 model did not experience degradation in classification accuracy for noise power level of 1\%, 10\%, 20\% and 40\%. Furthermore, the degradation in classification accuracy experience at noise power level of 60\% and 100\% are very minimal (1\% and 4\% respectively) as compared to the other two models. This is  unexpected as the shallowest model among the models under review is the VGG16 model and the deepest is the Shufflenet\_V2\_R1.0X model.  This may be because  the different architecture design philosophy which is the major difference among the models under review. The VGG16 and ResNet50 models were designed with the aim of maximizing the performance of deep neural network by leveraging on the ability of neural network to give improved performance with increase in the depth of the network. Although this design philosophy gave some consideration to the amount of compute , memory and power resource needed to train and test the models, maximizing performance was the primary consideration. This design philosophy is different from the design philosophy for the Shufflenet model where the same amount of consideration was given to both performance and the amount of memory, compute and power resource needed to train and test the models. This is primarily because Shufflenet was designed to be deployed on resource constrained devices with limited compute, memory and power budget.

The performance comparison of ResNet50 and ResNet50-SE when noise is injected into their layers is shown in Figure \ref{fig:ResNet50_se}. The performance of the individual models also follows the trend observed in section \ref{subsec:caseA} as expected.  Although they share similar names, ResNet50-SE is more resistant to noise as shown in Figure \ref{fig:ResNet50_se}. This performance difference can be attribute to slight difference in architecture design philosophy as ResNet50-SE is a variant of the original ResNet50.ResNet50 uses an identity-based skip connections to ensure deeper network are able to learn better. ResNet50-SE model, in addition to the identity based skip connection, uses Squeeze and excitation unit to ease the learning process, and significantly enhance the representational power of the network.it achieves these by explicitly modelling the inter-dependencies between the channels of its convolutional features  by using global information to selectively emphasise informative features and suppress less useful ones~\cite{resnet,Resnetse}.

\subsection{Analysis of models with similar design principles and depth but different number of parameters}
\label{subsec:caseC}
Model compression can be defined as any action taken to make DL models lighter or smaller in order to achieve faster inference or reduce their memory requirement~\cite{cheng,Han2016DeepCC}. The effect of model compression on performance of neural network models in the presence of additive white Gaussian noise in their layer is investigated in this section. This was achieved by using a model with similar architecture design principle and same depth but differ in the number of parameters. The models used for the investigation are the Shufflenet\_V2\_1x and Shufflenet\_V2\_0.5x \cite{shufflenet}. The Shufflenet\_V2\_1x model has 2.4 million parameters and 146 million floating point operations per second, while the Shufflenet\_V2\_0.5x model has 1.4 million parameters and 41 million floating point operations per second, highlighting the difference in their model complexity despite having the same depth.
 
\begin{figure}
  \includegraphics[height=5.85cm, width=8cm]{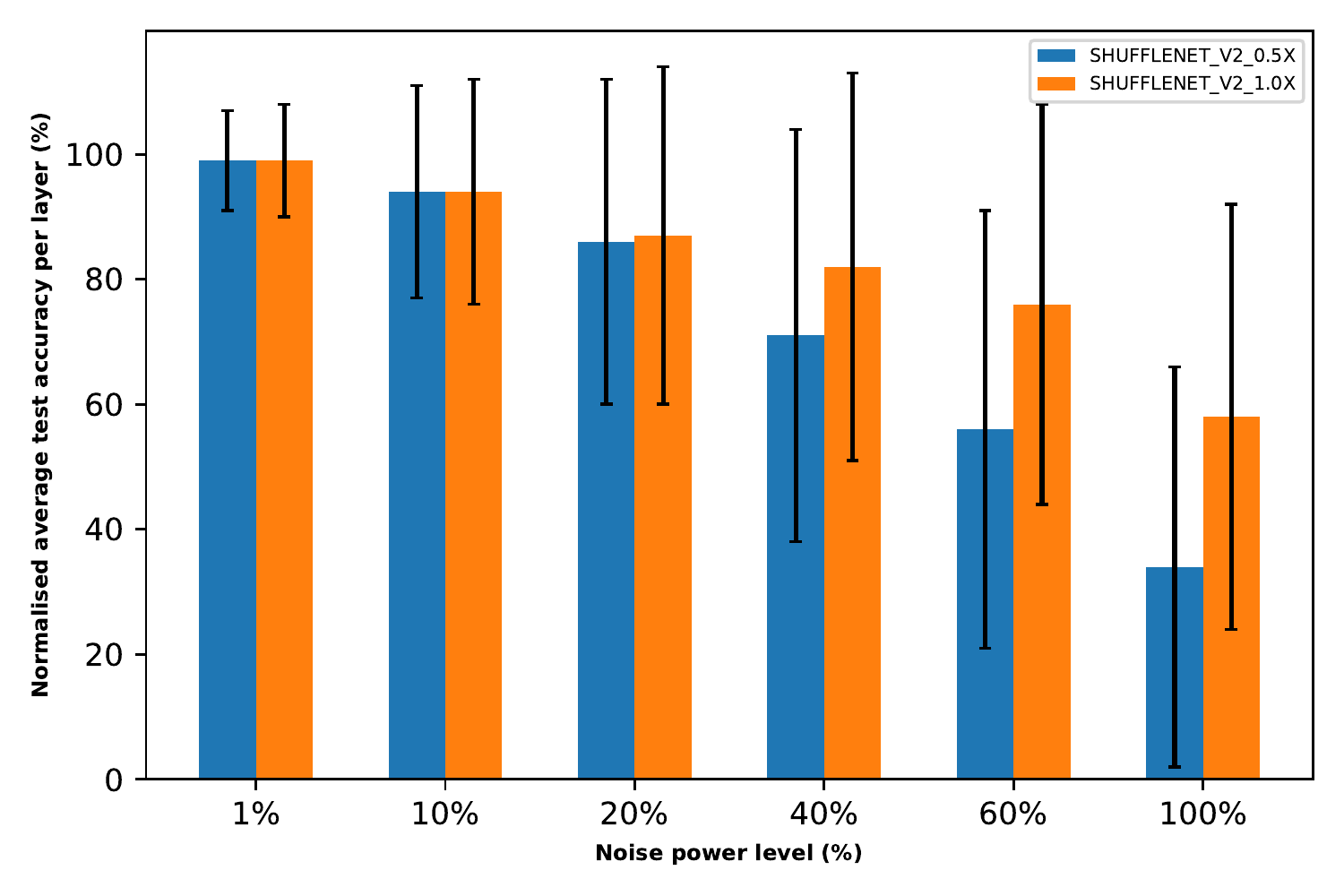}
\caption{Comparison of the performance of pre-trained Shufflenet\_V2\_1x and Shufflenet\_V2\_0.5x in the presence of noise in its layer during inference when tested with Imagenet dataset}
\label{fig:shufflenetcomparison}
\end{figure}

\begin{table}
\centering
\caption{Details of the Shufflenet\_V2\_0.5x and Shufflenet\_V2\_1X. The batch normalization parameters are grouped together and considered as a layer. The number of parameter includes both trainable and non-trainable parameters}
\label{tab:shufflenetcomparison}
\begin{tabular}{ | c | c | c | c |}
     \hline
     Model\_Name & \# of Layers & \# of Parameters \\ \hline \hline	
     Shufflenet\_V2\_0.5x  & 113 & 1,374,744\\ \hline
     Shufflenet\_V2\_1X & 113 & 2,295,940 \\ \hline
\end{tabular}
\end{table}

The performance of the two models are compared in Figure \ref{fig:shufflenetcomparison}. The performance of the individual model when additive white Gaussian noise of certain power level was injected into layers follow the trend established in section \ref{subsec:caseA}. However, there was a noticeable difference between the performances of the models despite having the same depth and following the same design principles. The Shufflenet\_V2\_0.5x perform poorly than the Shufflenet\_V2\_1x for most of the noise power levels. The poor performance is attributed to the difference in the model complexity of the models. The reduction in the number of parameters reduces the noise resistance ability of the Shufflenet\_V2\_0.5x model.

\subsection{Summary of observations}
In summary, the following observations can be made based on the experiments:
\begin{enumerate}
\item In general, deep models have some noise resistant ability. This ability was observed to be generally true irrespective of the model type and the dataset used. The noise resistant performance of the deep learning models decrease as the energy of the noise increases and/or the  number of layers of the model affected by the noise increases.
\item The noise resistant ability of various layers of deep learning models differ. It was observed that the layers that are deeper in the model are less sensitive to noise as compared to shallow layers in the model.
\item The number of layers in a model affects the noise resistant ability of a deep learning model. This explains the difference in performance of ResNet18, ResNet34 and ResNet50 models trained on Imagenet dataset in the presence of noise.
\item The architecture design principle of a deep learning model also affects its noise resistant ability. This explains the difference in performance between the VGG16, Shufflenet\_V2\_1X and ResNet50 models as shown in Figure \ref{fig:VGG16_SHUFF_ResNet}.
\item Model compression methods could impact on the noise resistance ability of a deep learning model. This was demonstrated by Shufflenet models as shown in Figure \ref{fig:shufflenetcomparison} that the noise resistant ability was impaired due to reduction in the number of parameters despite having the same depth.
\end{enumerate}

\section{{Discussion and Related Works}}
\label{sec:discussion}
There are several works in the literature on the  effect of analog noise on neural network models. The use of noise for regularization purpose of deep learning model was proposed in~\cite{Bishop}. It also considered 3 approaches to controlling trade-off between the bias and the variance of a neural network. The works of~\cite{Murray,Qin} focused on the injection of noise to the synaptic weight of neural networks during training to improve their noise resistant ability.The improvement in noise resistant ability is very important as analog devices suffers from limited precision and also presence of analog noise in the hardware. The use of knowledge distillation,a method to transfer knowledge from a teacher model to a student model, was used to improve further the noise resistant ability of a DL model in~\cite{NoisyNN}.In this case, the student and the teacher model are the same model. The student model obtained from this method showed better performance  than the  DL model obtained from the method used  in~\cite{Murray} during inferencing in the presence of analog noise.

Instead of training the DL models on digital devices, the training of neural networks on analog hardware in order to improve their noise resistant ability was done in~\cite{Bo,schmid}. The DL model is trained using stochastic gradient descent and back-propagation in the presence of noise it will later experience during testing. The resulting model adapt to the noise by developing some resistance to it in the process of learning its weight.Despite its advantages, this method has some demerits. The need to design a back-propagation circuitry on a device for only inferencing complicates the chip design and also increases the area of the chip though the back-propagation is only needed once. This method is laborious as DL training might have to be done on every chip that will be used during testing. This is because the analog noise power might vary from one chip to another~\cite{mixedsignal}. Chip-in-the-loop method, a method that involves measuring the error that the model is going to experience in the hardware and then using the error to update the weights in software, was used in~\cite{Bayraktaroglu,NeuroSmith}.This method is used to tune pre-trained model for the chip in order to make them adaptable to the chip. The process can be slow and very inefficient.  The use of linear and non-linear analog error correcting codes to protect the analog weights was done in~\cite{Upadhyaya2019ErrorCF}. A study on the design of code rate of error correction code for different layers of the neural network was also conducted. A comparison was done between this method and the binarized neural network which is an alternative to this method. The use of deep reinforcement learning and selective protection scheme to choose the important bit for error correcting codes protection was done in~\cite{huang2020functional}. The reinforcement learning algorithm was used to determine the complex relationship between the bits to protect and the model performance in order to determine the optimal trade off between redundancy and performance. The paper also showed that the most important bit in the weights of a DL model that is worthy of protection is not always  the most significant bit.

The existing works are different from this study as they explore ways to use analog noise to improve on the noise resistant properties of DL models. On the other hand, the objective of this study is to investigate and provide intuitions and insights about the noise resistant ability of popular deep neural networks, especially those DL models for image classification. Although some bench marking studies were done in~\cite{Upadhyaya2019ErrorCF}, digital noise (bit flipping) was used unlike this work where we used analog noise (additive white Gaussian noise). Furthermore, a comprehensive study has been performed in this work, including examining the performance of the DL models when the noise is added to all the weights, and the detailed effects of the noise on each individual layer of the DL models. The effect of noise on the layers of each neural network is important as layers are the building blocks of DL models. In addition, this study compared the performance differences between models of difference design and provide insights on the effect of model compression on the noise resistance ability of DL models.

\section{{Conclusion}}
\label{sec:conclusion}
Analog hardware implemented deep learning models are considered a promising approach for edge learning because they have faster execution speed and at the same time very energy efficient comparing to digital devices. However, one of the major concerns of such an approach is the analog noise incurred weights change in the deep learning models. In this paper, the noise resistance ability of various deep learning models in the presence of additive white Gaussian noise was investigated. Specifically, systematic experiments are carried out by adding white Gaussian noise of various power level to the weights in some of the popular deep learning models for image classification such as VGG, ResNet, and Shufflenet models. In addition to adding noise to all the weights of the models, experiments of adding noise to weights layer-by-layer are also conducted to examine the role of each layer in the deep learning models in terms of noise resistance. This has been done for deep learning models with different design principles, models with similar design principles but different depths or different number of parameters, as well as models with and without compression. A performance metric to assess the resistance of a model to the presence of noise in its layer was introduced and used to compare the performance of various models.

In this work, it was observed that deep learning models do have some noise resistant ability measured by their performance degradation due to various noise levels and it  varies from model to model. It was shown that the amount of degradation experience due to the presence of noise could be affected by factors such as number of parameters (model compression), depth of the model and design philosophy of the model. 
It was also observed that deeper layers in the models are more resistance to noise than the earlier layers.
This is an important step towards understanding how deep learning models would perform when implemented in analog hardware with noise affecting the performance of such models. The hope is that through comprehensive study, practitioners may be able to choose the appropriate deep learning models for given analog hardware to meet performance requirements. Furthermore, designers of deep learning models may find new ways to design the structure of the models and new methods to train the models that make them more robust to noise when implemented in analog hardware.

\section{Acknowledgments}
\label{acknowledgement}
This research work is supported in part by the U.S. Dept. of Navy under agreement number N00014-17-1-3062 and the U.S. Office of the Under Secretary of Defense for Research and Engineering (OUSD(R\&E)) under agreement number FA8750-15-2-0119. The U.S. Government is authorized to reproduce and distribute reprints for governmental purposes notwithstanding any copyright notation thereon. The views and conclusions contained herein are those of the authors and should not be interpreted as necessarily representing the official policies or endorsements, either expressed or implied, of the Dept. of Navy  or the Office of the Under Secretary of Defense for Research and Engineering (OUSD(R\&E)) or the U.S. Government.

\bibliographystyle{IEEEtran}

\bibliography{AutoencoderCNN}

$ $ \\

\begin{IEEEbiography}[{\includegraphics[width=1in,height=1.25in,clip,keepaspectratio]{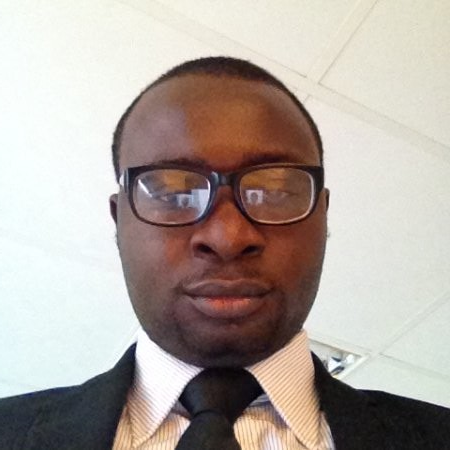}}]{Omobayode Fagbohungbe} is currently working towards his Ph.D. degree at U.S. DOD Center of Excellence in Research and Education for Big Military Data Intelligence (CREDIT Center),  Department of Electrical and Computer Engineering, Prairie View A\&M University, Texas, USA.  Prior to now, he received the B.S. degree in Electronic and Electrical Engineering from Obafemi Awolowo University, Ile-Ife, Nigeria and the M.S. degree in Control Engineering from the University of Manchester, Manchester, United Kingdom. His research interests are in the area of big data, data science, robust deep learning models and artificial intelligence.
\end{IEEEbiography}

%

\begin{IEEEbiography}[{\includegraphics[width=1in,height=1.25in,clip,keepaspectratio]{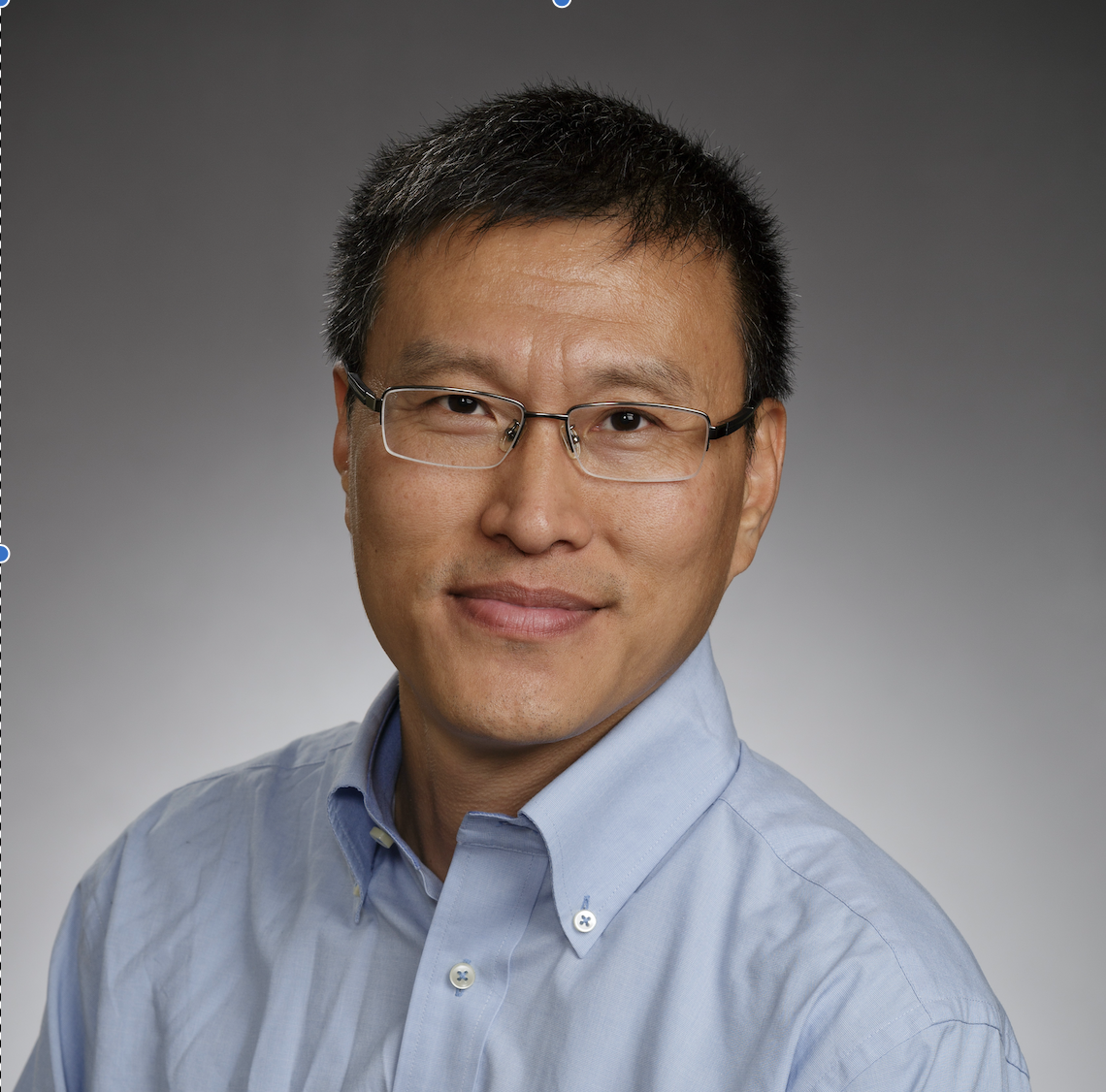}}]{Lijun Qian} (SM'08) is Regents Professor and holds the AT\&T Endowment in the Department of Electrical and Computer Engineering at Prairie View A\&M University (PVAMU), a member of the Texas A\&M University System, Prairie View, Texas, USA. He is also the Director of the Center of Excellence in Research and Education for Big Military Data Intelligence (CREDIT Center). He received BS from Tsinghua University, MS from Technion-Israel Institute of Technology, and PhD from Rutgers University. Before joining PVAMU, he was a member of technical staff of Bell-Labs Research at Murray Hill, New Jersey. He was a visiting professor of Aalto University, Finland. His research interests are in the area of big data processing, artificial intelligence, wireless communications and mobile networks, network security and intrusion detection, and computational and systems biology.
\end{IEEEbiography}


\end{document}